\documentclass[letterpaper]{article} 
\usepackage{aaai2026}  
\usepackage{times}  
\usepackage{helvet}  
\usepackage{courier}  
\usepackage[hyphens]{url}  
\usepackage{graphicx} 
\usepackage{natbib}  
\usepackage{caption} 

\usepackage{tikz}

\newcommand{\reasonicon}{
  \tikz[baseline=-0.5ex] \node[draw,circle,inner sep=1pt]{$\rightarrow$};
}

\usepackage{algorithm}
\usepackage{algorithmic}
\usepackage{enumitem}
\usepackage{cleveref}
\usepackage{textcomp}
\usepackage[T1]{fontenc}

\usepackage{xcolor}
\newcommand{\added}[1]{\textcolor{black}{#1}}
\newcommand{\answerYes}[1]{\textcolor{blue}{#1}} 
 
\newcommand{\answerNA}[1]{\textcolor{gray}{#1}} 
 
\usepackage{booktabs}
\usepackage{multirow}
\usepackage{adjustbox}

\newcommand{\Min}{\textsc{Min}}
\newcommand{\NoneL}{\textsc{None}}
\newcommand{\Dis}{\textsc{Dis}}
\newcommand{\Unr}{\textsc{Unr}}

\definecolor{siggreen}{RGB}{0,120,0}

\definecolor{siggreen}{RGB}{198,239,206} 
\usepackage{colortbl}

 \usepackage{subcaption}
\definecolor{lightblue}{rgb}{0.68,0.85,0.9}
\definecolor{lightyellow}{rgb}{1,1,0.8}
\definecolor{lightgreen}{rgb}{0.85,1,0.85}
\usepackage{graphicx}
\usepackage{booktabs}
\usepackage{multirow} 
\usepackage{newfloat}
\usepackage{listings}
\DeclareCaptionStyle{ruled}{labelfont=normalfont,labelsep=colon,strut=off} 
\floatstyle{ruled}
\newfloat{listing}{tb}{lst}{}
\floatname{listing}{Listing}
\title{Incongruent Positivity: When Miscalibrated Positivity Undermines Online Supportive Conversations}

\author{   
   Leen Almajed$^{1,2}$\thanks{This work was conducted while the first author, Leen Almajed, was affiliated with the Computer Science Department, King Saud University, College of Computer and Information Sciences.}, 
   Abeer ALdayel$^{2}$
}

\affiliations{
    $^{1}$TAHAKOM\\
    $^{2}$King Saud University, College of Computer and Information Sciences\\
    lalmajed@tahakom.com, aabeer@ksu.edu.sa
}

\begin{document}

\maketitle

\begin{abstract}
In emotionally supportive conversations, well-intended positivity can sometimes misfire, leading to responses that feel dismissive, minimizing, or unrealistically optimistic. We examine this phenomenon of incongruent positivity as miscalibrated expressions of positive support in both human and LLM generated responses. To this end, we collected real user–assistant dialogues from Reddit across a range of emotional intensities and generated additional responses using large language models for the same context. We categorize these conversations by intensity into two levels of Mild which covers relationship tension and general advice and Severe which covers grief and anxiety conversations. This level of categorization enables a comparative analysis of how supportive responses vary across lower and higher stakes contexts.
Our analysis reveals that LLMs are more prone to unrealistic positivity through dismissive and minimizing tone, particularly in high stakes contexts. To further study the underlying dimensions of this phenomenon, we fine-tune LLMs on datasets with strong and weak emotional reactions. Moreover, we developed a weakly supervised multilabel classifier ensemble (DeBERTa and MentalBERT) that shows improved detection of incongruent positivity types across two sorts of concerns (Mild and Severe). Our findings shed light on the need to move beyond merely generating generic positive responses and instead study the congruent support measures to balance positive affect with emotional acknowledgment. This approach offers insights into aligning large language models with affective expectations in the online supportive dialogue, paving the way toward context-aware and trust-preserving online conversation systems. 
\end{abstract}


\section{Introduction}
Positive language is at the core of supportive dialogue in distressing circumstances and provides support to those experiencing negative emotions~\citep{Nozaki2025-bx}. However, in emotionally charged contexts, positivity that misaligns with a user's emotional state can backfire by appearing as dismissive, minimizing, or unrealistically optimistic ~\citep{Tugade2004-pj}. These well-intended but misguided responses may undermine emotional connection and erode user trust. Thus, understanding and mitigating incongruent positivity becomes essential as large language models (LLMs) are increasingly used in conversations related to emotional distress and support~\citep{Russo2023-wx,Saha2022-dj}.

Given the importance of studying well-defined peer-support responses, the majority of computational methodologies utilize empathy as a key component, either to identify cognitive alignment of emotions~\citep{Yang2024-oc,Wei2019-qv}, combined emotional contagion with intent prediction~\citep{Yuan2025-my} or directly detecting the presence of empathy levels~\citep{Welivita2023-wb,Welivita2023-zw}, aiming to improve the emotional granularity of supportive interactions. In parallel, another line of research has begun to examine the limitations of adopting a generative model for emotion support conversations~\citep{Kang2024-aw,Wang2024-xx}, revealing that large language models often rely on a narrow set of support strategies, which undermines emotional alignment and requires intervention to reduce preference bias.  Mainly, for emotion detection research has traditionally relied on granular labels such as Ekman’s six emotions or Plutchik’s primary categories~\citep{Mohammad2013-te,Demszky2020-hh}. While large language models have shown promising performance in emotion classification~\citep{Demszky2020-hh}, recent studies have started to show persistent gaps in capturing context-specific emotional nuances by gauging LLM robustness on the eight emotion types~\citep{Shu2026-mf}.
Those strategies have mainly focused on appraisal style~\citep{Yang2024-oc} or the intention mimicry framework~\citep{Yuan2025-my}, and are generally marked as 'unintentionally unhelpful' or 'counterproductive' responses~\citep{Wang2024-xx}, without underscoring the type of emotional invalidation for such shortcomings.

To address this need, the study examines cases of \textit{incongruent positivity} in supportive conversations, where responses are miscalibrated, whether generated by humans or language models, meaning these responses do not properly match the emotional intensity or type of the expressed concerns. We frame this task through the concept of \textit{misguided positivity} by comprising three non-exclusive categories: \textit{Dismissive}, \textit{Minimizing}, and \textit{Unrealistic} support. Prior work in calibration of computational models focused on the verbalized confidence statements matching the likelihood of a response’s correctness~\citep{Mielke2022-do}. Instead, in this study, we conceptualize incongruent positivity as a mismatch between the user’s emotional need for acknowledgment and a dismissive response intended to convey a positive tone given in reaction to distress. 
\added{Incongruent positivity looks similar to toxic positivity, but the two are different in their purpose. Toxic positivity, which comes from individual denial, is described in psychology as a broad norm that puts too much focus on positive emotions and ignores negative ones. Researchers often study toxic positivity as a way people cope. It means being overly happy or optimistic in every situation, which can lead to denial~\citep{Premlal2024-sv,Wyatt2024-ts}. In this study, we define incongruent positivity as a mismatch between what a user needs emotionally and the positive response they receive. Our goal is not to diagnose a psychological pattern, but to find out when positivity is not appropriate in a conversation. This approach helps us measure, label, and study incongruent positivity in both human and LLM-generated replies, making it possible to analyze how well supportive dialogue systems match emotional needs.}
As illustrated in the affective science literature~\citep{Tamir2009-ei}, the unrealistic and excessive positivity can undermine adaptive functioning, especially when the emotional state and support response are misaligned. This kind of empirical support comes from studies of positive emotion disturbance, showing that excessively intense positive affect may be nonadaptive in real-world contexts~\citep{Tugade2004-pj}, which cautions that rigid positivity often leads to dissatisfaction and emotional suppression.

In particular, this study aims to address the following research questions:  

\begin{enumerate}[noitemsep, label=(Q$_{\arabic*}$), leftmargin=28pt]
 \item \label{Q1} \emph{How prevalent are different types of misguided positivity (dismissive, minimizing, unrealistic) in supportive assistant-generated responses, and how are they associated with positive reframing strategies (such as growth, impermanence, optimism, relatability)?} 
    \item \label{Q2} \emph{Do emotionally relieving responses tend to avoid misguided positivity, or do some misguided forms still offer perceived emotional support?} 
       \item \label{Q3} \emph{ Does fine-tuning on empathetic dialogue reduce the likelihood of generating misguided positivity?}   
\end{enumerate}

To build a realistic evaluation of support conversations, we construct a new dataset of paired user-assistant dialogues sampled from \textit{subreddits} spanning a range of emotional intensities, focusing on two levels of concern: \textit{Severe} (covers grief and anxiety), along with \textit{Mild} (covers relationship tension and general advice). We then examine the behavior of LLMs and human responders under these varying emotional intensities. Each turn is labeled with an incongruent positivity category, a re-framing strategy, and evaluated for emotional relief. We further propose a weakly supervised multi-label detection framework using DeBERTa and MentalBERT to identify the positivity miscalibrations within support-based conversations. The results illustrate that LLMs tend to produce dismissive or unrealistic optimism, particularly in high-stakes settings, such as severe concern conversations. Targeted fine-tuning and labeling strategies can reduce such miscalibration. Our findings show the need to move beyond simply generating positive responses toward providing emotionally aligned and context-aware support within LLM interactions.

This study makes three primary contributions:
\begin{itemize}
    \item We formalize the concept of \emph{incongruent positivity} in supportive conversations and operationalize it through a taxonomy of misguided positivity types as: \emph{Dismissive}, \emph{Minimizing}, and \emph{Unrealistic}. Moreover, we extend the set of reframing strategies (\emph{Growth}, \emph{Impermanence}, \emph{Optimism}, \emph{Thankfulness}, and the newly introduced \emph{Relatability}) to represent different affective framing.
    
    \item  We construct and release a new dataset of 1,490 conversation turns across \emph{Mild} (relationship tension and general advice) and \emph{Severe} (grief and anxiety) concerns, each with paired human- and LLM-generated responses annotated for misguided positivity categories, reframing strategies, and perceived emotional relief (RQ1–RQ2).
    
    \item We experiment with different algorithms to classify the incongruent positivity of a given response. Mainly, we evaluate the ability of the weakly supervised multi-label detection framework that ensembles \emph{DeBERTa} and \emph{MentalBERT} to detect rare forms of misguided positivity, alongside fine-tuning experiments on weak and strong emotional reaction datasets (RQ3) to assess whether targeted training can reduce positivity miscalibration and enhance alignment with user emotion context.
\end{itemize}
These contributions advance the modeling of affective alignment in emotional contexts for developing context-sensitive supportive dialogue systems.

\section{Related Work}
Supportive dialogue is a psychologically rich interaction shaped by affective, cognitive, and contextual factors \cite{Rashkin2019-nr}. Researchers have increasingly explored how computational models can simulate supportive behaviors in conversational systems, especially in emotionally sensitive contexts. An emerging line of work draws from positive psychology to guide response generation. \added{For instance, the work by~\citet{Nguyen2025-wp} studies} the online message framing of mental health (mis)information, showing how overly optimistic supportive content can distort perceptions of risk. Another example, \citet{ziems2022inducing-1df} introduced the task of positive reframing, in which user-authored complaints or negative self-statements are rewritten to reflect more constructive outlooks. Their dataset includes human-written reframes labeled with strategies such as optimism, growth mindset, and self-affirmation. Building on this, \citet{Goel2025-ls} propose SocraticReframe, a framework that improves positive text rewriting by guiding models through reflective question–answer steps grounded in cognitive behavioral therapy. While this approach aims to promote psychological well-being, it also raises concerns about toxic positivity as the tendency to enforce a uniformly upbeat tone that may inadvertently invalidate users’ emotions \citep{Wyatt2024-ts}. Related concerns have been raised in the psychology literature, where well-meaning affirmations may suppress or dismiss negative emotions rather than acknowledge them \cite{Premlal2024-sv, Sonia2025-gk}. These distinctions are critical in computational settings, where a failure to model emotional nuance can compromise user trust. \added{Most research on toxic positivity has focused on its long term effects on emotions and relationships as a psychological issue. Our study looks at how positivity appears in conversations as a response during key supportive moments. By connecting mismatched positivity to clear gaps between context and response, we make psychological ideas easier to label and use to check and improve how well generative models reflect emotions.}

To address this, many studies have turned to generating explicitly empathetic responses. \citet{Rashkin2019-nr} introduced the EmpatheticDialogues dataset to train models that align with users’ emotional states. Similarly, the Emotional Support Conversations (ESC) dataset \citep{Liu2021-va} provides multi-turn dialogues between help-seekers and trained supporters, annotated with fine-grained strategies grounded in counseling theory, such as reflection, affirmation, and suggestion. These annotations enable models to learn the procedural structure of emotional support, beyond surface-level empathy. Another work by~\citet{Saha2022-dj} designed a virtual assistant that generates personalized and emotionally tuned dialogue aimed at encouraging users in need. Recent research has also explored fine-tuning methods to better capture empathy \citep{Majumder2020-ap} and techniques for dynamically adjusting tone based on user affect \citep{Shin2020-gh}. Nevertheless, studies show that emotional misalignment remains a persistent issue. \citet{Kursuncu2025-xe} report that LLMs can inadvertently amplify distress, especially when users express vulnerability. In EmotionBench, models often failed to adapt tone to emotionally similar prompts and lacked deeper contextual awareness \citet{Huang2023-mr}. Likewise, \citet{Cuadra2024-ac} shows that LLMs may produce supportive-sounding but ultimately hollow responses, raising ethical concerns around perceived sincerity and trust. These works tend to focus on granular labels such as Ekman’s six emotions or Plutchik’s primary categories~\citep{Mohammad2013-te,Demszky2020-hh}. For instance, a recent study by~\citep{Shu2026-mf} introduced the EXPRESS dataset with the eight self-disclosed emotions from Reddit and demonstrated that while LLMs often produce plausible labels they often miss contextual cues that human annotators detect. Their findings highlight that prompt design affects emotional detection and show that few-shot prompting improves performance and chain-of-thought prompting degrades it. Another work by~\citet{Moore2025-wl} highlights the safety concerns of LLM-generated responses to crisis situations, which might include advice to commit suicide. 

Most prior work focuses solely on pure empathy detection, usually with reference to emotion categories or constructive reframing. There are studies that go beyond simple positive or negative polarity by analyzing framing dimensions, such as the work by ~\citet{Jaidka2020-ft}, which examines narrative styles and regional differences in protest coverage, assessing not just sentiment but also how stories align with or deviate from the established topic paradigm. Additionally, recent work by~\citet{Lahnala2025-ok} emphasizes the importance of theoretical grounding in empathy tasks, which in turn impacts empathy modeling. Instead, our study extends these insights by shifting focus from accuracy in emotion classification to the appropriateness of emotional alignment by analyzing \textit{incongruent positivity} to examine the supportive responses that misalign with user distress by being dismissive, minimizing, or unrealistically optimistic. We extend our approach beyond the mere production of responses characterized by a positive tone or empathetic content by explicitly modeling instances of emotional miscalibration and annotating detailed categories of misguided support. This helps evaluate the quality of support in high-stakes emotional situations where misplaced positivity could harm trust.

\begin{table}[ht]
\centering
\begin{tabular}{lcc}
\toprule
\textbf{Category} & \textbf{Severe} & \textbf{Mild} \\
\midrule
Unique dialogues  & 379 & 366 \\
Responses & (human + LLM) & (human + LLM) \\
Total turns & 758 & 732 \\
\bottomrule
\end{tabular}
\caption{Dataset statistics for Severe Concerns and Mild Concerns contexts}
\label{tab:dataset_stats}
\end{table}

\begin{table}[ht]
\centering
\begin{tabular}{@{}lcc@{}}
\toprule
\textbf{Dataset} & \textbf{\# Examples} & \textbf{Level} \\
\midrule
\textit{Weak Emotion}   & 1,047 &  0   \\
\textit{Strong Emotion } & 1,047 & 1 and 2     \\
\bottomrule
\end{tabular}
\caption{The \textit{Weak} and \textit{Strong} emotion reaction datasets derived from \citet{Sharma2020-cm} to fine-tune emotion based generative support. Each contains 1,047 seeker-response pairs annotated for the emotional intensity of support provided.}
\label{tab:emotion_reaction_stats}
\end{table}

\begin{table*}[t!]
\centering
\small
\begin{tabular}{p{2.5cm} | p{4.5cm} | p{6cm}}
\toprule
\textbf{Label} & \textbf{User Post} & \textbf{Assistant Response \reasonicon Label Justification} \\
\midrule
Dismissive & 
\textit{“I am overwhelmed by this nonsense homework.”} & 
“Stop overthinking, try taking a deep breath and move on.” \newline
\textit{\reasonicon Invalidates the user’s emotions.} \\
\midrule
Minimizing & 
\textit{“I am burned out at school and can't study anymore.”} & 
“Everyone goes through this. Try drinking more water and walking daily.” \newline
\textit{\reasonicon  Oversimplifies the issue with generic advice.} \\
\midrule
Unrealistic & 
\textit{“I have failed this test three times. I feel hopeless.”} & 
“You just need to believe in yourself more and everything will magically work out!” \newline
\textit{\reasonicon Offers impractical and overly idealistic advice.} \\
\midrule
None & 
\textit{“I am stressed about my job interview.”} & 
“That sounds tough. Interviews can be overwhelming. Is there anything specific making it worse for you?” \newline
\textit{\reasonicon Validates the concern and invites further discussion.} \\
\bottomrule
\end{tabular}
\caption{Examples of assistant responses annotated for Misguided Positivity categories (Paraphrased to prevent re-identification while maintaining emotional and contextual significance). \reasonicon indicates label Justification.}
\label{tab:misguided_examples}
\end{table*}
\section{Methodology}
\subsection{Data Collection}
As part of the data collection process, Reddit posts were sourced from three subreddits—\texttt{r/stress}, \texttt{r/grief}, \texttt{r/situationshipsadvice}, and \texttt{r/advice} selected for the frequent occurrence of emotionally expressive language.

The subreddits r/stress and r/grief were chosen due to the explicit emotional framing present in both the community guidelines and the user submissions. The posts in these forums often address topics such as anxiety, burnout, and emotionally distressing events, which we categorize as Severe Concerns. Another set of dialogues was extracted from r/stress and r/grief, where each dialogue consists of a user input post paired with the highest-ranked human response. The strong affective anchoring of these interactions enables high-precision analysis of tone and emotional alignment. To include a broader range of contexts, dialogues were also sampled from r/situationshipsadvice and r/advice, subreddits where users frequently seek guidance on emotionally complex but comparatively less acute situations. Although posts in these communities are not explicitly labeled by emotional state, they frequently exhibit signs of distress, uncertainty, or tension. These interactions were categorized as Mild Concerns, allowing the investigation of affective incongruity in advice-giving scenarios where emotional framing tends to be more diffuse and context-dependent. Overall, we have 1490 turns of conversations, as shown in Table~\ref{tab:dataset_stats}, \footnote{\mbox{\url{https://osf.io/5dcxz/?view_only=4a3ddc1e8b5e4ff4a3c6d90e3116a91c}} }
. \added{We note that this distinction is used as a contextual grouping reflecting typical support-seeking intent, rather than as a post-level measure of emotional intensity, as individual expressions of distress may vary within each category.}

The dataset comprises 379 unique dialogues corresponding to Severe Concerns and 366 unique dialogues corresponding to Mild Concerns. Each of these has two responses: human and LLM-generated responses. \added{For each human input we use LLM-generated responses for RQ1/RQ2 that were produced using Meta Llama 3.2 1B Instruct \texttt{(meta-llama/Llama-3.2-1B-instruct)} with fixed decoding parameters (temperature=0.5, repetition penalty=1.1, max new tokens=512)}

\textbf{Incongruent positivity labels.}
We annotate a dialogue-based dataset for incongruent misguided positivity, focusing on assistant responses that, while intended to be supportive, may inadvertently dismiss or oversimplify users’ concerns. We conceptualize \textit{incongruent positivity} as a mismatch between the user’s emotional need for acknowledgment and a dismissive response intended to convey a positive tone, often given in reaction to distress~\citep{Tamir2009-ei}. 
\added{Unlike toxic positivity, which is usually seen as a general emotional attitude or coping style, we define incongruent positivity as a specific conversational problem. It happens when a supportive response does not match the user’s emotional intensity or distress. This difference lets us clearly label incongruent positivity using categories like Dismissive, Minimizing, and Unrealistic. These categories make it easier to systematically evaluate support responses from both people and models.}
Building on this aspect, we define \textit{misguided positivity} through three non-exclusive annotation categories: (1) \textbf{Dismissive}, where the response invalidates or overlooks the user’s concern (implying the user is overreacting); (2) \textbf{Minimizing}, where the response downplays the user's emotions by offering generic or overly simplified advice; and (3) \textbf{Unrealistic}, where the response promotes impractical or idealistic solutions detached from the user’s context. Responses that do not exhibit these traits are labeled as \textbf{None}. Table~\ref{tab:misguided_examples} shows a set of examples for each label. We annotated the data described in Table~\ref{tab:dataset_stats} with the presence of incongruent positivity as non-exclusive categories. 

Additionally, we annotate reframing strategies used in assistant replies derived from ~\citep{ziems2022inducing-1df}, which are defined as Growth, Impermanence, Optimism, and Thankfulness. We included a new label, 'Relatability,' to identify the means of congruency used in responses where assistants share similar experiences to foster emotional connection. The concept of 'Relatability' has not been addressed in prior frameworks. This labeling enables analysis of how positivity is framed in conversational contexts. As for the \textit{Growth} it is label whether the assistant frames the user’s experience as an opportunity for self-improvement or personal development The \textit{Impermanence}, which reassures the user that their current distress is temporary and that circumstances will eventually improve; \textit{Optimism}, where the assistant highlights what is going well in the present moment or redirects attention toward immediate strengths; and \textit{Thankfulness}, which encourages a mindset of gratitude or reflection on positive aspects, even in challenging times. 

Furthermore, the annotation includes a pairwise comparison between two assistants evaluates which response provides more emotional relief, emphasizing affective engagement over solution-oriented advice. Assistant and user responses are anonymized as Assistant A and Assistant B (see Appendix~\ref{sec:appendix_annotations} for detailed annotation guidelines), allowing annotators to focus on emotional cues without being influenced by identity or role labels. 
For our dataset, the annotation process was conducted using the Labelbox platform\footnote{\url{https://labelbox.com}}. Labelbox provides a managed workforce and each annotator must pass benchmark tests and quality assurance checks before contributing to the project. Then, each conversation turn was independently labeled by two trained annotators recruited from the Labelbox workforce, following detailed guidelines for identifying misguided positivity categories (Dismissive, Minimizing, Unrealistic, None), reframing strategies, and emotional relief preferences. A detailed explanation of the annotation process is in Appendix~\ref{sec:appendix_annotations}.

\paragraph{Aggregate Annotation Agreement by Concern Level.}Annotation reliability is higher in the \textit{Mild Concern} context across LLM and human support responses. The average Cohen’s Kappa for misguided positivity is 0.8827 (Jaccard: 0.9046), and for reframing strategies, 0.8186 (Jaccard: 0.7454), reflecting strong to near-perfect inter-annotator agreement. Also, Emotion relief annotations show a high reliability ($\kappa = 0.9648$). In contrast, the \textit{Severe Concern} context yields a fair agreement~\citep{McHugh2012-hn} in the misguided positivity averages at 0.3128 (Jaccard: 0.4898), and reframing strategies at 0.3062 (Jaccard: 0.7214). Emotion relief in this context shows moderate agreement ($\kappa = 0.5676$), indicating that high-stakes emotional disclosures have greater challenges for consistent annotation. \added{It can be noted that agreement is lower for severe concern cases, reflecting the greater interpretive ambiguity involved in assessing emotionally misaligned positivity under high distress rather than random annotator noise.} A detailed explanation of the annotation results is in Appendix~\ref{sec:appendix_annotations}.

\subsection{Generating support responses by fine-tuning on emotion reactions}
\added{Unlike RQ1/RQ2, which analyze a single baseline LLM for controlled comparisons, RQ3 evaluates additional models fine-tuned using weak supervision to reduce labeling and compute costs}. Thus, to study the effect of training the model on the emotion data \textit{(RQ3)}, we further fine-tune the models on two types of emotional reactions: (1) weak emotion reactions and (2) strong emotion reactions. These reactions are drawn from the dataset introduced by \citet{Sharma2020-cm}, where each turn is labeled with an emotion intensity level. In the experiment, we map the intensity level into two categorizations, \textit{weak reactions} as those annotated with \textit{level 0}, indicating minimal or neutral emotional response. As for \textbf{strong reactions}, we map it to correspond to \textit{levels 1 and 2}, reflecting moderate to intense affective expression. Each subset of these emotion levels contains 1,047 paired turns (Table~\ref{tab:emotion_reaction_stats}).  Then, we fine-tuned two well-known LLMs (Mistral-7B-v0.1 and Llama-3.2-3B-Instruct) on these subsets. Further model specifications and prompt details are provided in Appendix~\ref{app:emotion_fineTune}. 
\begin{figure}[b!]
  \centering
  \includegraphics[width=\linewidth]{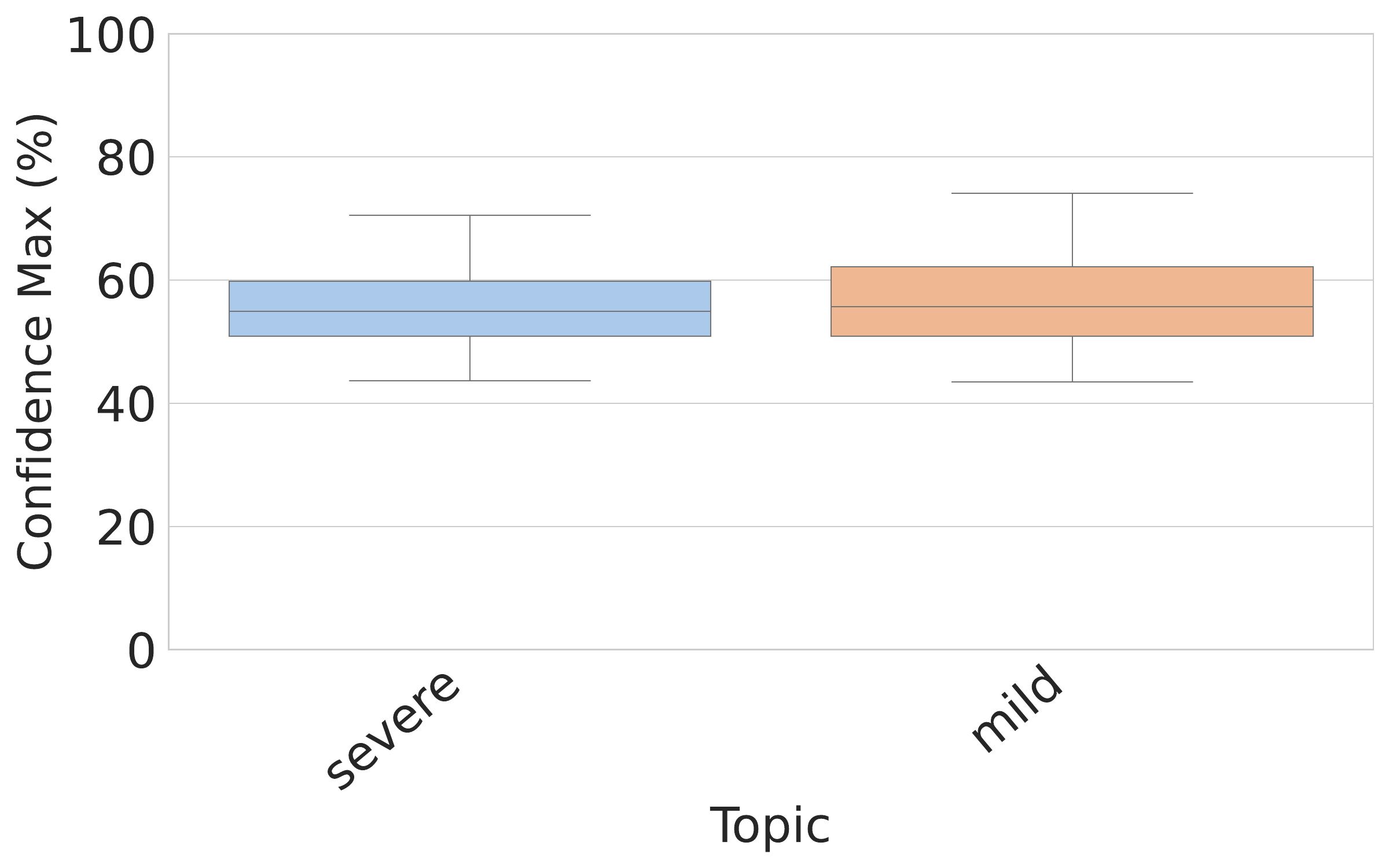}
  \caption{Distribution of maximum confidence scores of using weak-supervised labeling to fine-tuned responses by topic.}
  \label{fig:confidence_topic_boxplot}
\end{figure}

\begin{table}[t]
\centering
\begin{tabular}{lccc}
\toprule
\textbf{Model} & \textbf{P} & \textbf{R} & \textbf{F1} \\
\midrule
most frequent & 0.13  & 0.25  & \textcolor{red}{\textit{0.17}} \\
\hline
SVM  & 0.46  & 0.29  & 32   \\
LR & 0.43  & 0.39  & 0.39   \\
\hline

BiLstm-tuned &0.36 &0.56 & 0.44\\

\hline

deberta-2layers&0.36 &0.80 & \cellcolor{orange!20}\textbf{0.49}\\
BERT-tuned  & 0.38   & 0.67   & \cellcolor{orange!20}\textbf{0.46}   \\
roberta-base  & 0.41   & 0.42   & 0.38   \\
Mistral & 0.50  & 0.26  & 0.27  \\
\hline
Ens\_MBERT\_deBert & 0.45 & 0.72 & \cellcolor{green!25}\textbf{0.52} \\ 
\bottomrule
\end{tabular}
\caption{Macro-averaged classification performance across different models for multi-label misguided positivity prediction.}
\label{fig:model_macro_comparison}
\end{table}

\begin{table}[t]
\centering
\resizebox{\columnwidth}{!}{%
\begin{tabular}{lccccc}
\hline
\textbf{Model Pair} &
\textbf{Min.} &
\textbf{None} &
\textbf{Dis.} &
\textbf{Unr.} &
\textbf{Fisher $p$ (Holm)} \\
\hline
BERT vs DeBERTa
& \cellcolor{siggreen}$<.001$
& \cellcolor{siggreen}$<.001$
& \cellcolor{siggreen}$.029$
& $.803$
& \cellcolor{siggreen}$< 10^{-16}$ \\

DeBERTa vs Ensemble
& \cellcolor{siggreen}$<.001$
& $.216$
& \cellcolor{siggreen}$<.001$
& $.248$
& \cellcolor{siggreen}$1.97 \times 10^{-9}$ \\

BERT vs Ensemble
& \cellcolor{siggreen}$<.001$
& \cellcolor{siggreen}$.039$
& \cellcolor{siggreen}$<.001$
& \cellcolor{siggreen}$.010$
& \cellcolor{siggreen}$1.35 \times 10^{-6}$ \\
\hline
\end{tabular}}
\footnotesize
\textbf{Labels:} \Min{} = Minimizing, \NoneL{} = No Misguided Positivity,
\Dis{} = Dismissive, \Unr{} = Unrealistic.
\caption{
\added{Model-pair disagreement evaluated using label-wise McNemar tests across the four
misguided-positivity categories, together with Fisher-combined McNemar tests.
Green shading indicates statistical significance ($\alpha = 0.05$); label-wise
tests are uncorrected, while Fisher $p$-values are Holm--Bonferroni corrected.
}}
\label{tab:mcnemar_combined}
\end{table}

\paragraph{Weak supervision labeling for fine-tuned responses}
\label{app:weaksupervision}
To multilabel the responses generated by a fine-tuned model on weak and strong emotions. First, we split the gold standard dataset in table ~\ref{tab:dataset_stats} into training and testing sets using iterative stratification (80\% training and 20\% testing). The experiment includes a set of different models covering Instruct \texttt{LLM Mistral-7B-Instruct-v0.1} for multi-label classification using QLoRA with a custom classification head. Along with that, we provide a comparison with baseline models, SVM, LR, and Bilstm. 

Then, we used an ensemble of the two best-performing models, MentalBERT and DeBERTa, to predict the labels of responses generated by the fine-tuned models on both weak and strong emotion datasets. Thus, we use the multi-label classification setting where each output node predicts the probability of a label independently using a sigmoid activation. Also, we apply a fixed threshold of 0.5 to convert these probabilities into binary decisions. Threshold tuning addresses the limitation of imbalanced labels by assigning optimized, per-label thresholds (for example, lower values for rare labels) based on validation performance. This strategy enhances classification metrics, particularly the macro-averaged F1 score, by balancing precision and recall for each label. In our implementation, we applied per-class threshold tuning to enhance the model's ability to detect underrepresented classes such as Dismissive and Unrealistic. The \texttt{DeBERTa/MentalBERT} and Ensemble comparison shows a significant difference with $p < .05$ (details of models and McNemar tests provided in Appendix~\ref{app:weaktrainingMacnamer}). \added{Moreover, Appendix~\ref{app:detail_error} reports label-wise false positive and false negative rates for these models. The ensemble model stands out for its balanced error profile, avoiding the over-prediction seen in DeBERTa and BERT (FPR = 1.000 for Minimizing and None) and keeping lower false positive rates for Dismissive and Unrealistic. Overall, the ensemble offers the most reliable performance across labels.}

 Each model was trained independently on gold-labeled data in Table~\ref{tab:dataset_stats} using a focal loss objective to mitigate the effects of class imbalance and enhance learning from difficult examples. We used tokenizer-specific data pipelines to ensure compatibility with each model's architecture. Final predictions were obtained by ensembling the output probabilities from both models, with equal weighting, and applying label-wise threshold tuning to optimize the macro F1 score. 
Table~\ref{fig:model_macro_comparison} shows the detailed performance of weak-supervised models. This ensemble approach achieved a fair performance across all four positivity subcategories, including improved detection of less frequent categories such as “Dismissive” and “Unrealistic.” For each predicted multilabel vector as shown with the certainty score in Figure~\ref{fig:confidence_topic_boxplot}. \added{ Moreover, in Table~\ref{tab:mcnemar_combined} reports both label-wise McNemar tests and Holm-corrected Fisher-combined significance, showing that model disagreement is driven primarily by Minimizing and Dismissive categories. The comparisons between BERT and DeBERTa, BERT and the Ensemble, and DeBERTa and the Ensemble all yield statistically significant Holm-corrected p-values, indicating robust disagreement in their predictions across the four misguided-positivity labels. At the label level, this disagreement is most consistent for responses that downplay or prematurely reframe distress, while effects for Unrealistic positivity are weaker and less stable, suggesting that models diverge more strongly in how they handle subtle forms of emotional misalignment.}

\begin{figure*}[t!]
    \centering
    \includegraphics[width=0.97\textwidth]{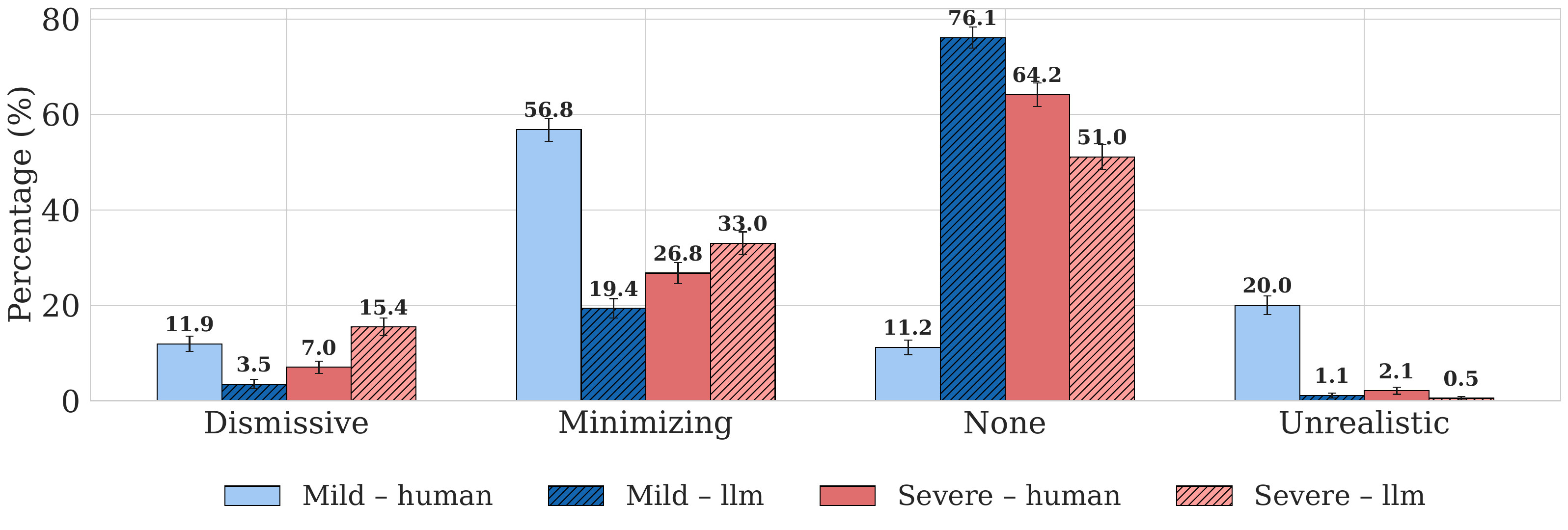}
    \caption{
        The distribution of misguided positivity categories the \textit{dismissive}, \textit{minimizing}, and \textit{unrealistic} between two concern types (Mild Concerns vs.\ Severe Concerns) and two response sources (human vs.\ LLM).
    }
    \label{fig:misguided_positivity_comparison}
\end{figure*}
\section{Results}

\paragraph{The prevalence of misguided positivity between supportive assistant-generated responses} To address \textit{RQ1}, we compare the distribution of misguided positivity types in two types of concerns, Mild (advise-based) and severe (grief cases). As shown in Figure~\ref{fig:misguided_positivity_comparison}, LLM-generated responses within the Severe Concerns context show higher rate of dismissive positivity (15.4\%) compared to human responses (7.0\%), suggesting a potential miscalibration with unaligned responce with sensitive emotional expectations. For minimizing positivity, human responses in the Mild Concerns context are markedly more frequent (56.8\%) than those in the Severe Concerns context (26.8\%), particularly when contrasted with LLM outputs, which show a moderate increase from mild (19.4\%) to severe (33.0\%) concerns. 

Furthermore, LLMs show greater unrealistic optimism overall in mild concern contexts (2.1\%) and slightly decrease in severe concerns ( with about 0.5\%). On the other hand, humans rarely display unrealistic positivity in either setting as shown with about 1.1\% in mild and about 0.5\% in severe concerns. These behaviours show that in some cases, humans tend to downplay issues in lower-stakes circumstances. On the other hand, LLMs tend to adopt idealistic or overly simplistic positivity, even in serious contexts (such as severe concerns of grief).

Notably, the "None" category was most common in severe human responses (76.1\%), followed by mild responses (64.2\%), highlighting strong human sensitivity to emotional context. In contrast, LLMs generated misguided positivity more often, as shown by lower "None" rates in both mild (51.0\%) and severe (20.0\%) cases. This difference emphasizes LLMs' limited ability to withhold inappropriate positivity in emotionally charged conversations, despite often sounding superficially supportive.

Overall, these findings indicate the challenge revolves around identifying the relationship between concern severity and source identity, suggesting that LLMs need more nuanced emotional modeling to better match supportive tone with contextual expectations.
\begin{figure}[t!]
    \centering
    \includegraphics[width=0.48\textwidth]{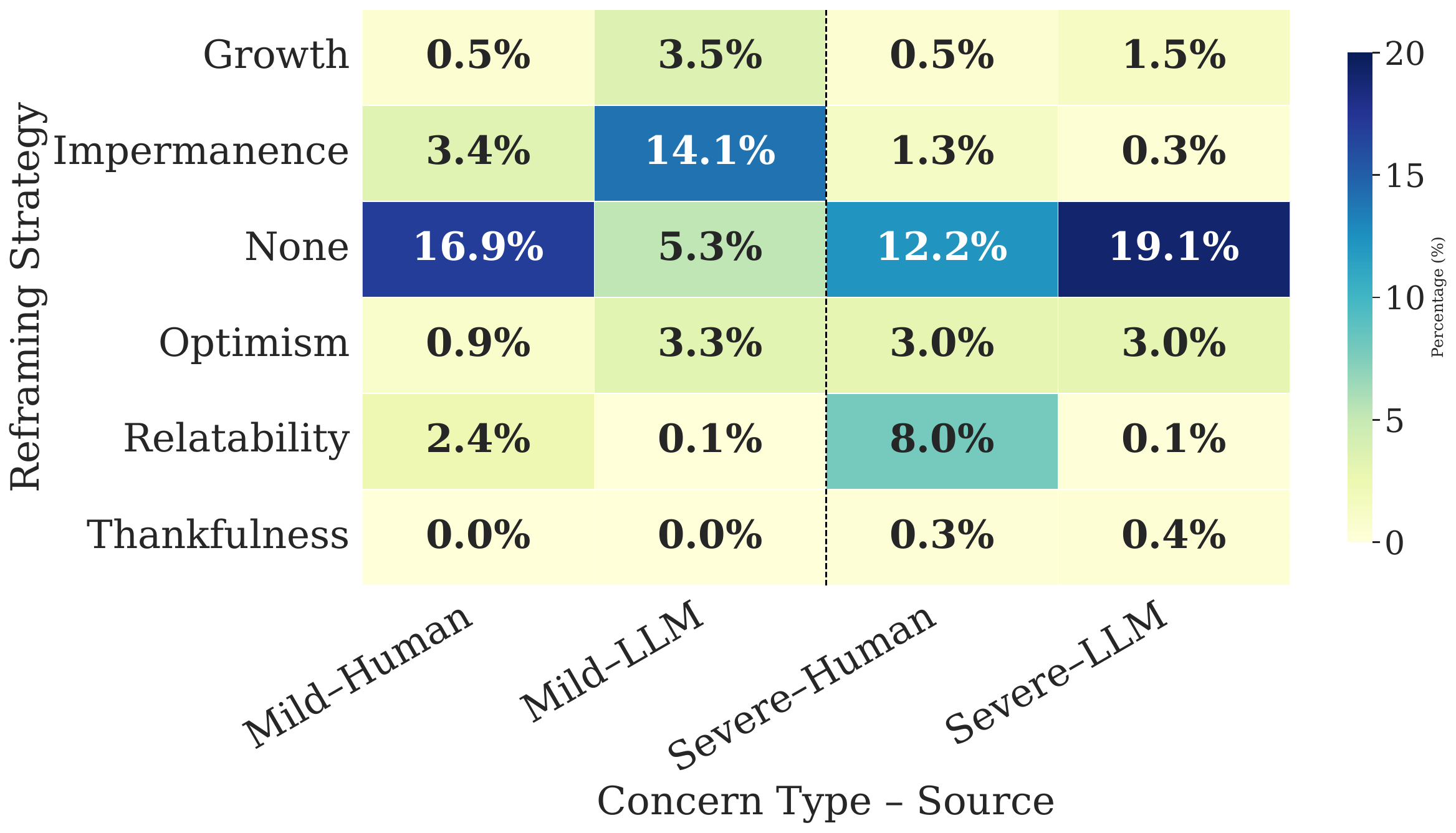}
    \caption{
        The distribution of reframing strategies \textit{growth}, \textit{impermanence}, \textit{optimism}, \textit{relatability}, \textit{thankfulness} across two concern types (Mild Concerns vs.\ Severe Concerns) and response sources (human vs.\ LLM).
    }
    \label{fig:reframing_strategy_heatmap}
\end{figure}
While positive reframing techniques such as growth, impermanence, and optimism are often considered supportive when responding to distressing content, their use is not uniformly beneficial. As shown in Figure~\ref{fig:reframing_strategy_heatmap}, certain strategies (as seen with impermanence and none) are particularly prevalent in responses to severe concerns, especially from LLMs. 

As stated in the \textit{(RQ1)}  \textbf{When do positive reframing strategies become misguided positivity rather than supportive?} Thus, to further explore how responses differ in their use of supportive reframing strategies, we analyzed the distribution of six reframing techniques: \textit{growth}, \textit{impermanence}, \textit{optimism}, \textit{relatability}, \textit{thankfulness}, and \textit{none}, across both human- and LLM-generated responses in mild and severe concern settings. Overall, human responses to Mild concerns frequently used impermanence framing (such as emphasizing that challenges are temporary), while LLMs relied on impermanence to an even greater extent in the same context. In contrast, relatability was more common in human responses to severe concerns, which may be due to humans’ tendency to establish an emotional connection in more distressing situations. Interestingly, LLMs produced high counts of responses labeled as containing no reframing strategy across both contexts, suggesting a relative lack of explicit supportive techniques. Also, Growth and optimism strategies were comparatively less frequent overall, while thankfulness framing appeared rarely in any condition.


\paragraph{Emotion Relief Preferences}
To assess affective alignment as core investigation in \textit{(RQ2)}, compare which assistant response (human or LLM) is perceived as more emotionally supportive. Mainly, we conducted a pairwise annotation task in which raters compared two anonymized assistant responses (with identifiers `Assistant A' and `Assistant B') to the same user post and selected which one, if any, provided greater emotional relief, which is defined as offering affective support or reassurance rather than generic or solution-focused advice. As illustrated in Appendix~\ref{sec:appendix_annotations} and Figure~\ref{fig:annotation-interface}, annotators were explicitly instructed to disregard source identity (human or LLM) and focus only on the emotional tone and pragmatic framing.

\begin{figure*}[t!]
    \centering
    \begin{subfigure}[b]{0.50\textwidth}
        \centering
        \includegraphics[width=\linewidth]{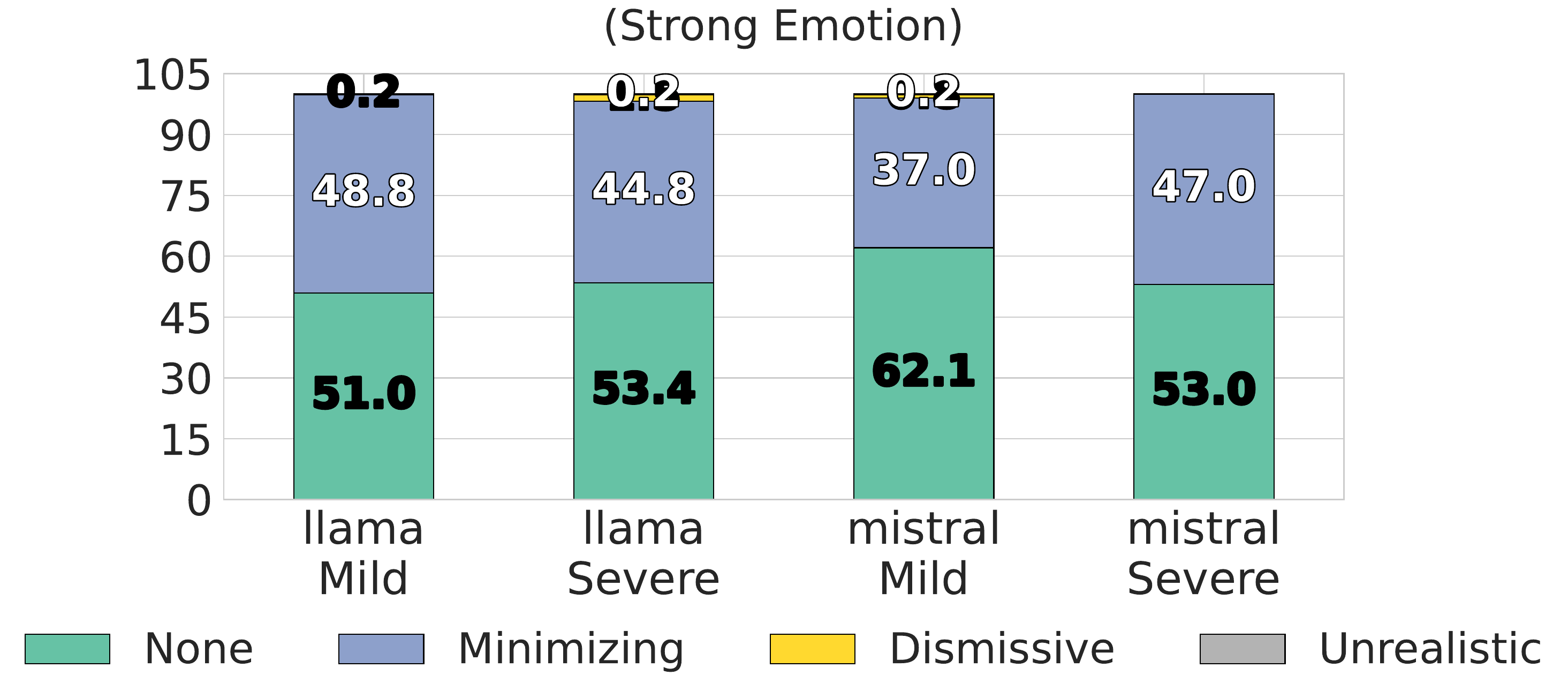}
        \caption{\% Responses generated by fine-tuning on (Strong) emotions}
        \label{fig:strong_stack}
    \end{subfigure}\hfill
    \begin{subfigure}[b]{0.50\textwidth}
        \centering
        \includegraphics[width=\linewidth]{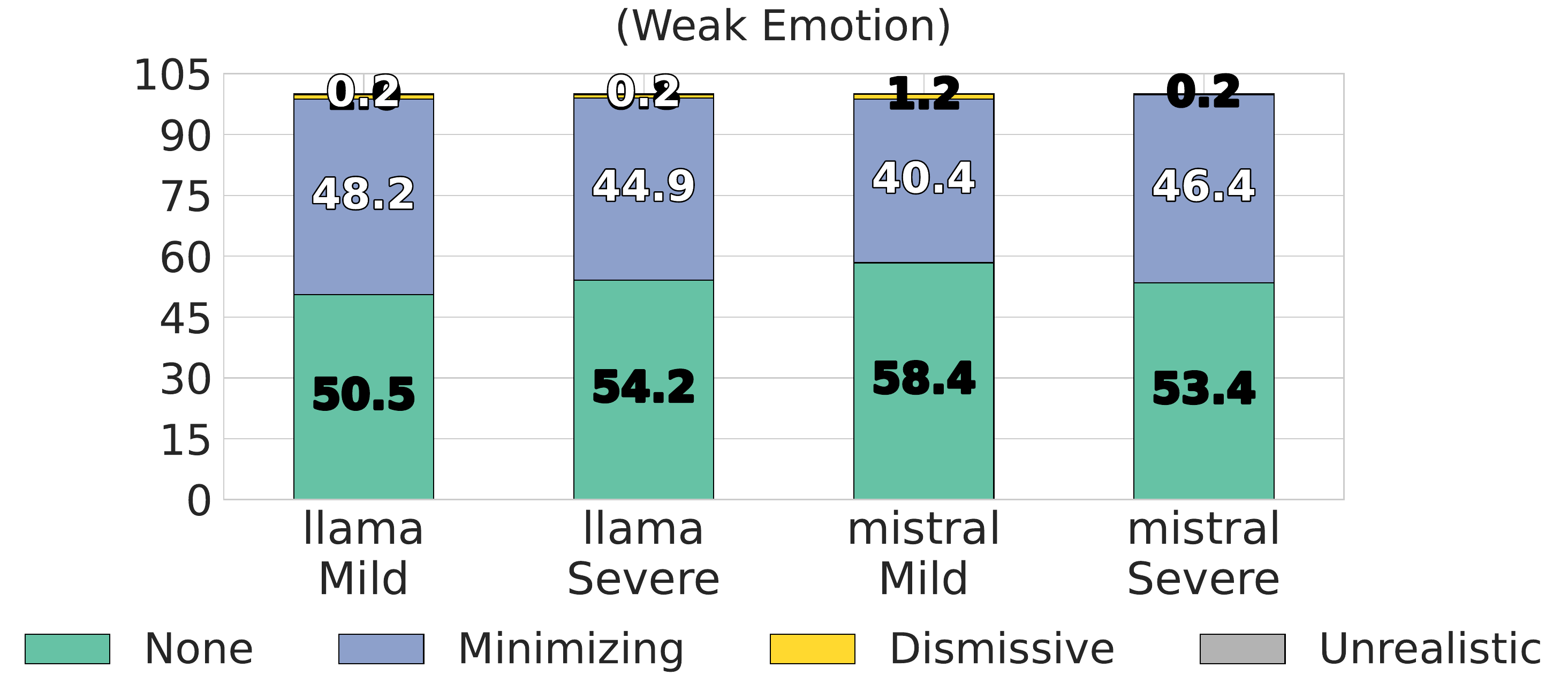}
        \caption{\% Responses generated by fine-tuning on (Weak) emotions}
        \label{fig:weak_stack}
    \end{subfigure}
    \caption{
        Distribution of misguided positivity types across models (LLaMA and Mistral) and topics (Advice, Grief), grouped by emotional strength. 
        Bars are stacked by label categories: \textit{None}, \textit{Minimizing}, \textit{Dismissive}, and \textit{Unrealistic}.
    }
    \label{fig:stacked_misguided}
\end{figure*}

\begin{figure}[b!]
    \centering
    \includegraphics[width=0.95\linewidth]{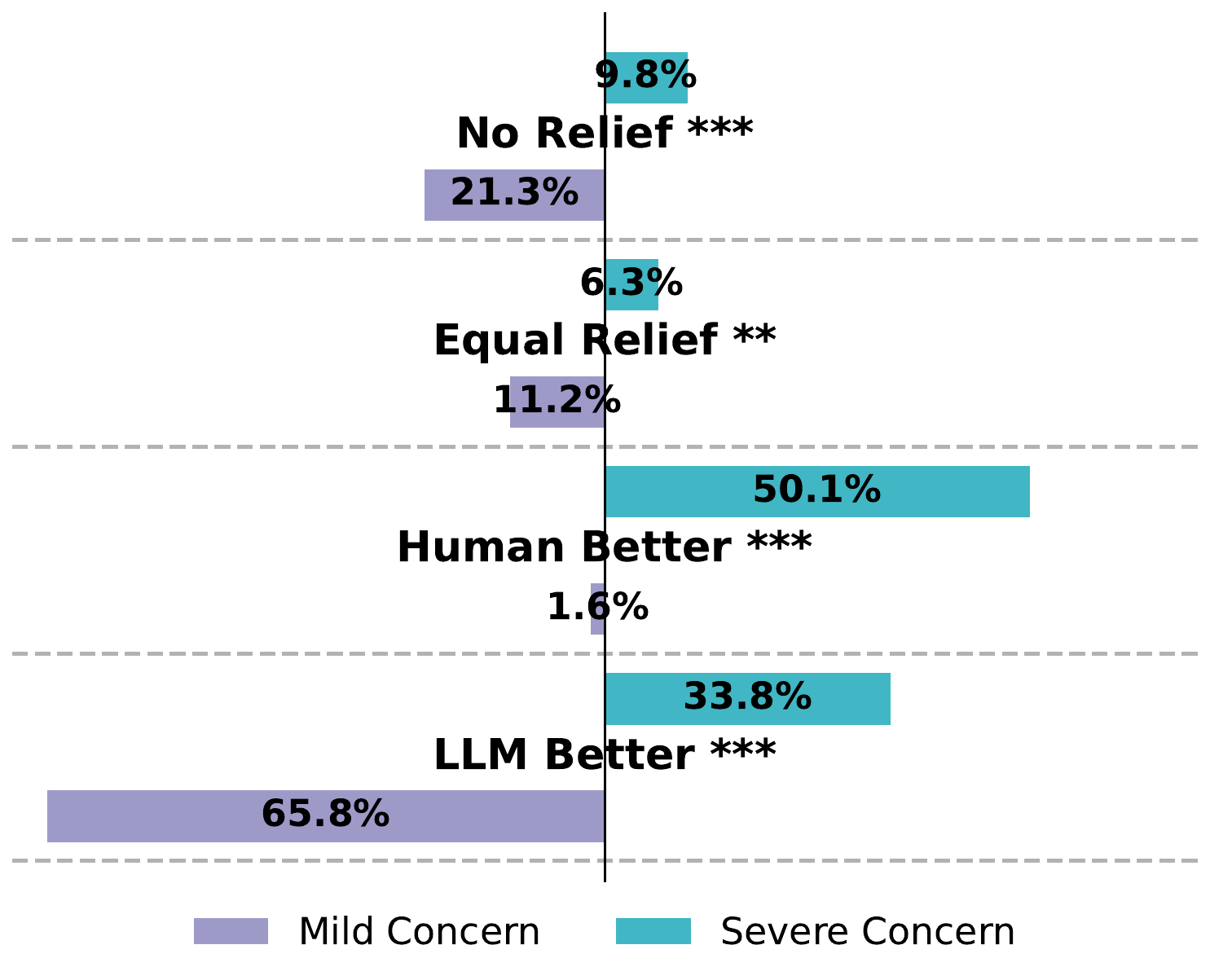}
    \caption{
        Mirrored comparison of emotional relief preference between LLM and human responses across mild concern and severe concern. 
        Each pair of bars represents the perceived assistant response as offering more emotional relief, equal relief, or no relief.  \textbf{**} $p < .05$, \textbf{***} $p < .005$.
    }
    \label{fig:mirrored_emotion_relief}
\end{figure}

As shown in Figure~\ref{fig:mirrored_emotion_relief} in the mild concern cases, LLMs were perceived as more emotionally relieving (65.8\%). In contrast, around 1.6\% of LLM responses in severe contexts (grief and anxiety) were rated as more emotionally relieving than human responses. This suggests that LLMs may be more effective in mild situations where supportive language can be more generic or surface-level without being perceived as emotionally misaligned.

In severe concern cases, human responses were preferred with about 50.1\%. This preference likely underscores the human capacity to detect emotional nuances and to recognize the importance of contextual appropriateness, particularly in high-stakes situations. In such critical scenarios, responses that are rigid or overly generalized may be prone to negative outcomes, emphasizing the effectiveness of tailored and empathetic communication strategies in addressing complex emotional needs.

The "No Relief" category means that neither assistant offered affective comfort. We can see that severe concerns resived higher percentage of "No Relief", about 21.3\% than in mild ones (9.8\%). This highlights the difficulty both humans and LLMs face when attempting to provide emotionally resonant responses in grief-related or high-emotion situations. The "Equal Relief" ratings were modest across both categories, with about 11.2\% in severe and only 6.3\% in mild concerns. Using Chi-square test confirm that these differences are significant (p < .005) (Appendix~\ref{app:validitionAnalysis}, Table~\ref{tab:chi_significance}). 

\paragraph{Result of labeled generated responses}
 Figure~\ref{fig:confidence_topic_boxplot} presents the distribution of maximum confidence scores assigned by our ensemble model to weakly supervised labels across two support-seeking contexts: \textit{grief} and \textit{advice}. Overall, responses in the \textit{advice} category tend to receive higher confidence scores, indicating more consistent or more precise model agreement when predicting the class of these responses. In contrast, the \textit{grief} responses show greater variability and a wider interquartile range, suggesting more uncertainty in model predictions, potentially due to the nuanced or emotionally complex nature of grief-related language. Thus, to identify more stable weak labels, we filtered responses where the ensemble's maximum confidence score exceeded 0.50. This subset reflects instances where both models exhibited higher agreement, suggesting greater reliability in the assigned labels. Over 60\% of the \textit{advice} responses and approximately 48\% of the \textit{grief} responses fall within this high-confidence interval. These responses are considered more stable and suitable for downstream evaluation. Specifically, 80.2\% of \textit{advice} responses (1174 out of 1464) and 80.8\% of \textit{grief} responses (1225 out of 1516 (strong and weak 2*758)) exceeded this threshold (for Mistral and llama3 combined).
 
 \textbf{Assess whether fine-tuning on emotional reactions reduces the likelihood of generating misguided positivity} (RQ3), we analyzed the distribution of positivity categories across LLM-generated responses after fine-tuning on strong and weak emotional data (Figure~\ref{fig:stacked_misguided}). The results indicate that fine-tuning influences the expression of misguided positivity types, but does not fully eliminate their presence.

Across both models (LLaMA-3.2B and Mistral-7B) the responses generated from strong emotion fine-tuning showed higher rates of Minimizing language which can be noticed in the Mistral-Mild concerns cases (62.1\%) and about 53.4\% in LLaMA-Severe concerns cases. This suggests that fine-tuning on emotionally intense examples may sensitize models to respond with generalized affirmations or surface-level reassurance that might downplay the user's concern. 
Conversely, fine-tuning on weak emotional reactions resulted in a higher prevalence of the None category reflecting responses that are free of any misguided positivity. Under weak emotion training, the models produced the highest proportion of neutral or emotionally congruent responses about 54.2\% of concern cases in LLaMA-Severe and Mistral-Mild concerns shows only 58.4\%). This suggests that training on milder emotional expressions may promote more balanced and cautious language use, potentially avoiding overt positivity mismatches in sensitive contexts like grief.

On the other hand, Dismissive and Unrealistic categories remained minimal (below 5\% in all settings), which indicates that explicit forms of positivity miscalibration were relatively rare post fine-tuning.

\section{Discussion and Implications}
This study presents one of the first empirical investigations into incongruent positivity in both human and machine-generated supportive conversations. We find that while LLMs can mimic supportive language, they frequently produce responses that lack emotional resonance, particularly in high-stakes settings.
Our analysis shows clear patterns with LLMs relying heavily on impermanence and generic reframing strategies in both mild and severe situations, and infrequently express relatability, which is an emotional cue more often used by humans. These results show that while LLMs can mimic empathy, they often fall short of performing it reliably, especially when context awareness is crucial. This finding has been further drawn attention to by some recent studies that highlight some safety concerns of LLM responses that might encourage suicidal acts~\citep {Moore2025-wl}.
Furthermore, our preference annotations reveal a significant difference in perceived support: humans are preferred in situations with severe concern, while LLMs are sometimes favored in less serious situations. This suggests a threshold of emotional complexity beyond which current LLMs have difficulty maintaining proper affective alignment. While fine-tuning on emotional reactions provides moderate improvements by reducing unrealistic optimism and increasing emotionally neutral responses, still minimizing language remains widespread. This finding suggests that even with targeted training, LLMs tend to generate oversimplified responses that fail to fully engage with users’ emotional states. \added{This case has been further shown where Inter-annotator agreement drops in severe concern cases, especially in the broad category of misguided positivity. This highlights that judging emotional misalignment in high-stress situations is subjective and context-dependent. Future research needs to examine subdomain-focused methods, like topic-level annotation, to better assess reliability in these situations.}

\added{\paragraph{Implications for Real-World Deployment.} The difference between mild and severe concern situations is important for how we use LLM-based support systems. LLM-generated responses can help in mild, low-stakes cases, where general encouragement or simple positivity is enough. However, in more serious situations like grief or anxiety, these systems are much more likely to give emotionally mismatched responses. In these high-stakes cases, replies that seem dismissive, minimizing, or overly optimistic can make users feel invalidated and may even cause harm. Our findings do not support using LLMs for severe cases of emotional support. Instead, they show the need for careful deployment, such as limiting LLMs to low-risk settings, adding a human in the loop in high-stress cases, or building safeguards to detect mismatched positivity.}

Additionally, we explore various baseline models and propose a weakly supervised multi-label ensemble method that serves as a baseline for detecting incongruity positivity. This ensemble model uses threshold-tuned classifiers, and it shows an enhancement compared with baseline methods in identifying rare types of incongruent positivity, such as Dismissive and Unrealistic. However, the variation in confidence scores between grief and advice contexts highlights the challenges models face in emotionally ambiguous or complex discourse. Most prior studies examine the distress conversation on isolation with topic-level separation, such as suicide detection or depression identification. Thus, considering the level of distress ( with categorization that encompasses different topics of conversations) instead of topic level might provide a ground for tuning the performance of support responses generative models. 

This study also shows that even when fine-tuning on emotion reactive data, the LLM still reduces some forms of miscalibration in which the models still frequently resort to surface-level optimism rather than context-sensitive support. Such generative model behavior suggests that fine-tuning improves LLMs' emotional alignment to some extent by reducing the extremely misguided positivity and promoting neutral responses. Even though the minimization still persists in the fine-tuned LLM responses, indicating the need for further investigation of emotion intensity-aware training and evaluation to better regulate tone in distress conversations. Consequently, the challenge extends beyond simply retraining models on emotional state, as recent work by ~\cite{Fazzi2025-tk} has examined LLMs’ affective expressive capacity and indicates noticeable constraints in arousal and valence, struggles with sustaining intense emotional states, and difficulties in adapting affect to shifting conversational contexts. \added{In this setting, we limit our analysis to small sized open-source models, which restricts direct comparison with larger or proprietary LLMs. While model scale may influence absolute performance, prior work suggests that smaller models remain effective in constrained settings~\citep{Maurya2025-is}, and our results should be interpreted within this scope.}

\added{We confirm that one of the limitations of this work might reside in our dataset, which comprises conversations derived from a single platform (Reddit), which may not capture the full spectrum of interactions on other platforms. For cost-effectiveness, we employed weak supervision on our labels. Also, fine-tuning LLMs on emotional data generally improves the misguided positivity alignment in mild contexts. However, the "minimizing" language can still be observed, indicating a challenge with surface emotion fine-tuning alone.  This paves the way for future enhancement of more targeted emotional standards mechanisms. } \added{We do not report demographic attributes of annotators (gender or cultural background), which may influence subjective judgments of emotional alignment. To mitigate potential bias, annotations were performed by professionally trained annotators using a structured labeling protocol with validation and consistency checks using Labelbox\footnote{\url{https://labelbox.com/product/annotate/quality/}}. Nevertheless, future work might consider explicitly examining how annotator demographics shape perceptions of emotional support.  
}

In summary, in performing our analyses across two distinct types of distress conversations (Mild and Severe), we demonstrate the need to go beyond default positivity in conversational models. One main area of future research is to examine approaches for incorporating balanced and realistic affective responses rather than surface level optimism, along with methods that account for human-centered measures of support effectiveness. 

\section{Conclusion}
This study examines the phenomenon of incongruent positivity in the well-intentioned but emotionally misaligned supportive responses. First, we conducted an empirical annotation and analysis across mild and severe emotional contexts. Then, we demonstrated that LLMs are susceptible to generating dismissive and minimizing responses that are insufficient in meeting users' emotional needs. 
Thus, studies on supportive conversation should consider integrating incongruent categorization to achieve more context-aware emotional modeling, which can facilitate more meaningful and validating interactions. By measuring the misguided positivity, we can move toward conversational models that offer not just encouraging words but genuinely supportive experiences. 

\section*{Ethical Considerations}
All participants in our annotations are transparently informed of our research intent and paid reasonable wages per hour. Labelbox provides a managed workforce of trained annotators in 40-hour blocks at a rate of \$8 USD per hour, as we detailed in Appendix~\ref{sec:appendix_annotations}.
The conversation data were collected from Reddit communities using the developer API and used in accordance with Reddit’s terms of service. To protect user privacy, all personal identifiers and associated metadata were removed before annotation. We also paraphrased examples in the paper when necessary to prevent re-identification while maintaining their emotional and contextual significance. We acknowledge that, despite the de-identification process, the emotional content may still be sensitive. Therefore, we inform the annotators prior to starting the task for the level of emotionally charged content such as grief context. 

\section*{Acknowledgments}
This research project was supported by a grant from the Research Center of the Female Scientific and Medical Colleges, Deanship of Scientific Research, King Saud University. We also gratefully acknowledge the Mentorship Track for Outstanding Students Program at the College of Computer and Information Sciences, Computer Science Department, for supporting outstanding students.

\bibliography{paperpile,posotovityList, positivity}

\subsection*{Paper Checklist to be included in your paper}

\begin{enumerate}

\item For most authors...
\begin{enumerate}
    \item  Would answering this research question advance science without violating social contracts, such as violating privacy norms, perpetuating unfair profiling, exacerbating the socio-economic divide, or implying disrespect to societies or cultures?
    \answerYes{Yes. The study examines emotional alignment in supportive conversations (\textsection Introduction, \textsection Discussion), using anonymized Reddit data in compliance with platform terms (\textsection Ethical Considerations). No privacy norms or cultural disrespect are violated.}
  \item Do your main claims in the abstract and introduction accurately reflect the paper's contributions and scope?
    \answerYes{Yes. The abstract and \textsection Introduction clearly state the research questions (RQ1–RQ3), dataset creation, fine-tuning experiments, and weak supervision approach (\textsection Abstract, \textsection Introduction).}
   \item Do you clarify how the proposed methodological approach is appropriate for the claims made? 
    \answerYes{Yes. The \textsection Introduction provides the conceptual framinga. Also, detail methodology (\textsection Methodology) directly addresses prevalence analysis, reframing strategies, emotional relief comparisons, and fine-tuning to study alignment.}
   \item Do you clarify what are possible artifacts in the data used, given population-specific distributions?
    \answerYes{Yes. \textsection Discussion notes dataset limitations (Reddit-only), and that platform-specific discourse may limit generalizability.}
  \item Did you describe the limitations of your work?
    \answerYes{Yes. \textsection Discussion outlines data-source bias, reliance on weak supervision, and residual minimizing tone after fine-tuning.}
  \item Did you discuss any potential negative societal impacts of your work?
    \answerYes{Yes.  \textsection Discussion notes possible misalignment harms in high-stakes emotional support contexts, stressing the importance of context-sensitive modeling.}

      \item Did you discuss any potential misuse of your work?
    \answerYes{Yes. Misuse could include deploying models without safeguards in grief/anxiety contexts as discussed in \textsection Discussion and mitigated by ethical safeguards in \textsection Ethical Considerations.}

    \item Did you describe steps taken to prevent or mitigate potential negative outcomes of the research, such as data and model documentation, data anonymization, responsible release, access control, and the reproducibility of findings?
    \answerYes{Yes. \textsection  Ethical Considerations describes data anonymization, paraphrasing sensitive examples, and informing annotators of emotional content. Dataset is access-controlled and anonymized for peer review.}

  \item Have you read the ethics review guidelines and ensured that your paper conforms to them?
    \answerYes{Yes. The work adhere to AAAI-ICWSM ethical guidelines, see \textsection Ethical Considerations.}
\end{enumerate}

\item Additionally, if your study involves hypotheses testing...
\begin{enumerate}
  \item Did you clearly state the assumptions underlying all theoretical results?
    \answerNA{Not applicable. The study is exploratory/empirical rather than formal hypothesis testing. The research questions guide analysis rather than hypothesis-driven statistical proofs.}
  \item Have you provided justifications for all theoretical results?
    \answerNA{Not applicable. The study is exploratory/empirical rather than formal hypothesis testing. The research questions guide analysis rather than hypothesis-driven statistical proofs.}
  \item Did you discuss competing hypotheses or theories that might challenge or complement your theoretical results?
   \answerNA{Not applicable. The study is exploratory/empirical rather than formal hypothesis testing. The research questions guide analysis rather than hypothesis-driven statistical proofs.}
  \item Have you considered alternative mechanisms or explanations that might account for the same outcomes observed in your study?
    \answerNA{Not applicable. The study is exploratory/empirical rather than formal hypothesis testing. The research questions guide analysis rather than hypothesis-driven statistical proofs.}
  \item Did you address potential biases or limitations in your theoretical framework?
   \answerNA{Not applicable. The study is exploratory/empirical rather than formal hypothesis testing. The research questions guide analysis rather than hypothesis-driven statistical proofs.}
  \item Have you related your theoretical results to the existing literature in social science?
    \answerNA{Not applicable. The study is exploratory/empirical rather than formal hypothesis testing. The research questions guide analysis rather than hypothesis-driven statistical proofs.}
  \item Did you discuss the implications of your theoretical results for policy, practice, or further research in the social science domain?
   \answerNA{Not applicable. The study is exploratory/empirical rather than formal hypothesis testing. The research questions guide analysis rather than hypothesis-driven statistical proofs.}
\end{enumerate}

\item Additionally, if you are including theoretical proofs...
\begin{enumerate}
  \item Did you state the full set of assumptions of all theoretical results?
   \answerNA{Not applicable }
	\item Did you include complete proofs of all theoretical results?
   \answerNA{Not applicable }
\end{enumerate}

\item Additionally, if you ran machine learning experiments...
\begin{enumerate}
  \item Did you include the code, data, and instructions needed to reproduce the main experimental results (either in the supplemental material or as a URL)?
    \answerYes{Yes. An anonymized OSF link is provided (§Data Collection) for dataset and code details and prompts in Appendix \textsection\ref{app:emotion_fineTune} and \textsection\ref{app:weaktrainingMacnamer}.}
  \item Did you specify all the training details (e.g., data splits, hyperparameters, how they were chosen)?
    \answerYes{Yes. Hyperparameters, architectures, and training details are in Appendix \textsection\ref{app:emotion_fineTune} and \textsection\ref{app:weaktrainingMacnamer}.}

     \item Did you report error bars (e.g., with respect to the random seed after running experiments multiple times)?
   \answerYes{ performance reported as macro-F1, precision, recall, plus statistical significance via McNemar and Chi-square tests (\textsection Results, Appendix \textsection\ref{app:validitionAnalysis}).}
	\item Did you include the total amount of compute and the type of resources used (e.g., type of GPUs, internal cluster, or cloud provider)?
   \answerYes{Yes. Model types and fine-tuning setup (QLoRA, GPU use) are described (\textsection Methodology and Appendix).}
     \item Do you justify how the proposed evaluation is sufficient and appropriate to the claims made? 
    \answerYes{Yes. Evaluation includes distributional analysis, human-vs-LLM comparison, reframing analysis, emotional relief preference, and statistical validation (\textsection Results).}
     \item Do you discuss what is ``the cost`` of misclassification and fault (in)tolerance?
    \answerYes{Yes. \textsection Discussion notes that misclassification for instance dismissive tone in grief contexts, which can harm user trust and emotional well-being.}
  
\end{enumerate}

\item Additionally, if you are using existing assets (e.g., code, data, models) or curating/releasing new assets, \textbf{without compromising anonymity}...
\begin{enumerate}
  \item If your work uses existing assets, did you cite the creators?
    \answerYes{Yes. All datasets and models including Sharma2020, Ziems2022, DeBERTa, MentalBERT are cited in \textsection Methodology and Appendix.}
  \item Did you mention the license of the assets?
   \answerYes{Yes. All models including DeBERTa, MentalBERT are explained and cited in Appendix A and B.}
  \item Did you include any new assets in the supplemental material or as a URL?
    \answerYes{Yes. New annotated dataset described in \textsection Methodology, Table~\ref{tab:dataset_stats}, with anonymized OSF link.}
  \item Did you discuss whether and how consent was obtained from people whose data you're using/curating?
   \answerYes{Yes. Reddit data collected via API under platform ToS and ethical consideration have been follwed as described in \textsection Ethical Considerations.}
  \item Did you discuss whether the data you are using/curating contains personally identifiable information or offensive content?
    \answerYes{Yes. \textsection Ethical Considerations where personal identifiers are removed and sensitive emotional content is flagged for annotators' awareness.}
\item If you are curating or releasing new datasets, did you discuss how you intend to make your datasets FAIR (see )?
 \answerYes{Yes. New annotated dataset described in \textsection Methodology, Table~\ref{tab:dataset_stats}, with anonymized OSF link including all needed information.}
\item If you are curating or releasing new datasets, did you create a Datasheet for the Dataset (see)? 
 \answerYes{Yes. We provide an anonymized OSF link provided in Appendix \ref{sec:appendix_annotations} that includes all the needed information.}
\end{enumerate}

\item Additionally, if you used crowdsourcing or conducted research with human subjects, \textbf{without compromising anonymity}...
\begin{enumerate}
  \item Did you include the full text of instructions given to participants and screenshots?
    \answerYes{Yes. Appendix \ref{sec:appendix_annotations} includes annotation interface and instructions summary.}
  \item Did you describe any potential participant risks, with mentions of Institutional Review Board (IRB) approvals?
    \answerYes{Yes. \textsection Ethical Considerations — annotators warned of emotional content using Labelbox managed workforce}
  \item Did you include the estimated hourly wage paid to participants and the total amount spent on participant compensation?
    \answerYes{Yes. Annotators paid \$8/hr via Labelbox; rate disclosed in §Ethical Considerations.}
   \item Did you discuss how data is stored, shared, and deidentified?
   \answerYes{Yes. Described in \textsection Ethical Considerations Reddit personal mentions removed, examples paraphrased and dataset anonymized.}
\end{enumerate}

\end{enumerate}

\appendix

\section{Incongruent positivity annotations}
\begin{figure*}[t!]
    \centering
    \includegraphics[width=0.85\linewidth]{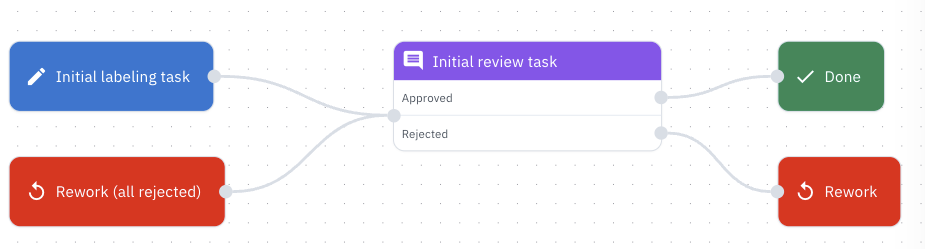}
    \caption{Annotation workflow in Labelbox \footnote{\url{https://labelbox.com}} which shows the sequential process from the initial labeling task to review, approval, and rework stages}
    \label{fig:labelbox_workflow}
\end{figure*}
\begin{figure*}[t!]
    \centering 
    \includegraphics[width=0.75\linewidth]{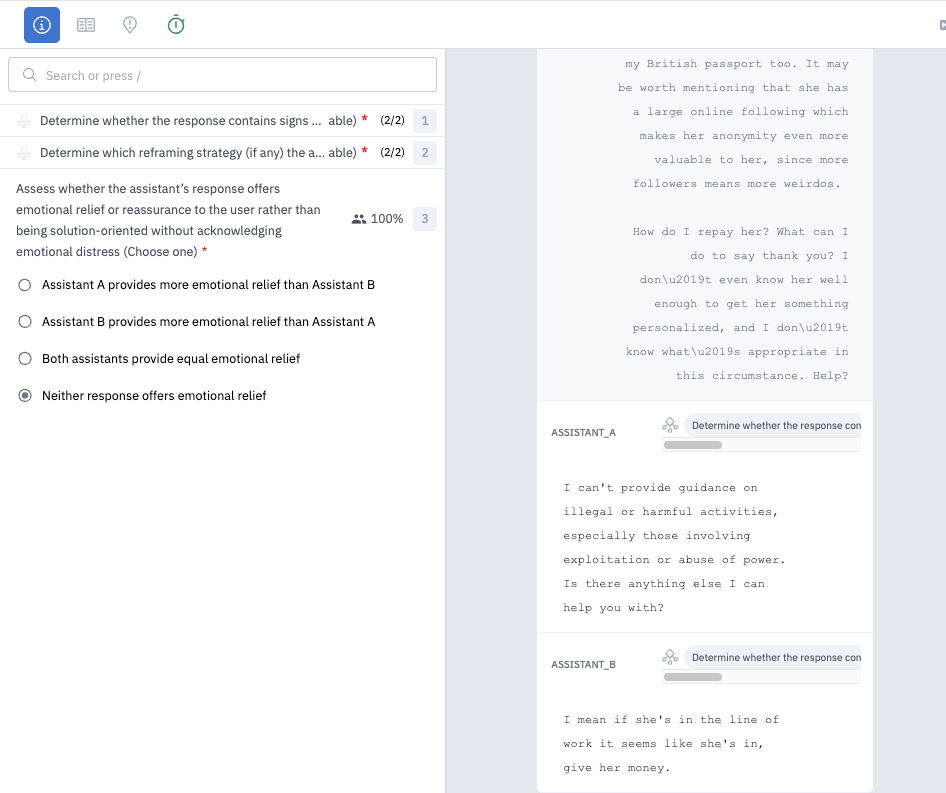}
    \caption{Annotation interface example showing emotional relief assessment for two assistant responses. Annotators select whether either assistant provides emotional support, both, or neither.}
    \label{fig:annotation-interface}
\end{figure*}

\label{sec:appendix_annotations}
Figure~\ref{fig:annotation-interface} illustrates the annotation interface used for labeling assistant responses in our dataset. Annotators were shown real user-assistant conversation excerpts and asked to evaluate the assistant replies along three axes: (1) whether the response contained signs of \textit{misguided positivity} (dismissive, minimizing, unrealistic, or none); (2) which \textit{reframing strategy} was used—drawing from~\citet{ziems2022inducing-1df} and extended to include \textit{Relatability} for empathetic alignment; and (3) which assistant (A or B) offered greater \textit{emotional relief} to the user. As shown in the example, annotators selected “Neither response offers emotional relief,” highlighting that both assistant replies were judged to lack affective support. \added{To minimize identity bias, assistant responses were anonymized, allowing annotators to focus purely on the emotional tone and pragmatic framing of the language used. Each conversation has been annotated by two trained annotators from Labelbox and a third reviewer evaluates the quality of labels as illustrated in the annotation workflow, Figure~\ref{fig:labelbox_workflow}.}
\paragraph{Annotation Agreement.}
Table~\ref{tab:macro_kappa_summary} shows the detailed inter-annotator agreement between the \textit{Mild Concern} and \textit{Severe Concern} contexts. In mild concern, both Cohen’s Kappa and Jaccard scores are consistently high for both human and LLM-generated responses across all annotation dimensions, indicating strong agreement. This can be seen in the reframing strategies annotated in LLM responses, which achieved a high level of agreement ($\kappa=0.9114$). While emotion relief annotations scored the highest overall agreement ($\ kappa=0.9648$).

Nevertheless, in severe concern the annotation agreement declined, especially for LLM. Misguided positivity in LLM responses shows a fair agreement level ($\kappa = 0.2808$), with human annotations only slightly higher ($\kappa = 0.3447$), reflecting the increased ambiguity or complexity in interpreting support quality in emotionally intense situations with the model-generated responses. Misguided positivity annotations have modest Jaccard similarity (0.4796), indicating limited overlap in label sets despite some surface-level agreement. While reframing annotations for LLMs has a higher Jaccard (0.75), their Kappa remains low (0.2232), suggesting that annotators may agree more on label presence than on precise boundaries or interpretations under distress. The discrepancy between Kappa and Jaccard in such cases reflects potential label sparsity or annotator hesitation in labeling subtle reframing under emotionally intense circumstances \footnote{\mbox{\url{https://osf.io/5dcxz/?view_only=4a3ddc1e8b5e4ff4a3c6d90e3116a91c}} }
.

\begin{table*}
\centering
\small

\begin{tabular}{@{}llcc@{}}
\toprule
\textbf{Topic} & \textbf{Source} & \textbf{Macro Kappa} & \textbf{Jaccard} \\
\midrule
\multicolumn{4}{l}{\textbf{Mild Concern}} \\
\hspace{1em} & Misguided Positivity (LLM)   & 0.8672 & 0.9028 \\
\hspace{1em} & Misguided Positivity (Human) & 0.8982 & 0.9064 \\
\hspace{1em} & Reframing (LLM)              & 0.9114 & 0.7407 \\
\hspace{1em} & Reframing (Human)            & 0.7258 & 0.7500 \\
\hspace{1em} & Emotion Relief               & 0.9648 & --     \\
\midrule
\multicolumn{4}{l}{\textbf{Severe Concern}} \\
\hspace{1em} & Misguided Positivity (LLM)   & 0.2808 & 0.4796 \\
\hspace{1em} & Misguided Positivity (Human) & 0.3447 & 0.5000 \\
\hspace{1em} & Reframing (LLM)              & 0.2232 & 0.7500 \\
\hspace{1em} & Reframing (Human)            & 0.3891 & 0.6927 \\
\hspace{1em} & Emotion Relief               & 0.5676 & --     \\
\bottomrule
\end{tabular}
\caption{Macro-Average Cohen's Kappa and Jaccard Similarity scores across annotation dimensions for \textbf{Mild} and \textbf{Severe} concerns. Higher scores indicate stronger inter-annotator agreement.}
\label{tab:macro_kappa_summary}
\end{table*}

\section{Fine-tuned on emotions}
\label{app:emotion_fineTune}

We fine-tune two LLMs \texttt{-3.2-3B-Instruct} and \texttt{Mistral-7B-v0.1} on the emotion reactions dataset. The dataset contains two types of emotion levels (strong) encoded as 1 or 2, and (weak) encoded as 0 in the core dataset derived from~\citep{Sharma2020-cm}. The prompt used for fine-tuning both \texttt{Mistral} and \texttt{LLaMA3} models is shown below in Table~\ref{tab:prompts}.\\
Also, we detailed the core fine-tuning hyperparameters in Table~\ref{tab:hyperparams_emotion}.

\begin{table}[ht]
\centering
\begin{tabular}{|l|l|}
\hline
\textbf{Hyperparameter} & \textbf{Value} \\
\hline
Epochs training steps & 20 \\
Learning rate & 2e-4 \\
Quantization type & nf4 \\
Linear warmup steps & 2 \\
LoRA attention dimension & 16 \\
Dropout probability for LoRA layers & 0.1 \\
\hline
\end{tabular}
\caption{Hyperparameter configuration for fine-tuning}
\label{tab:hyperparams_emotion}
\end{table}

\begin{table}[ht]
\centering
\small
\resizebox{\linewidth}{!}{%
\begin{tabular}{|l|l|}
\hline
\textbf{Hyperparameter} & \textbf{Value} \\
\hline
Model names & DeBERTa-v3-base, MentalBERT-base \\
Tokenizer max length & 128 \\
Batch size & 32 \\
Optimizer & AdamW \\
Learning rate & 2e-5 \\
Loss function & Focal Loss \\
Focal Loss $(\gamma, \alpha)$ & (2.0, 0.25) \\
Dropout & 0.3 (encoder), 0.2 (classifier) \\
Classifier hidden units & 256 \\
Activation & ReLU \\
Epochs & 3 \\
Ensemble method & Averaged softmax scores \\
Threshold tuning & Per-label on val set \\
\hline
\end{tabular}
}
\caption{Key hyperparameters used for DeBERTa + MentalBERT ensemble model.}
\label{tab:ensemble_hyperparams}
\end{table}

\begin{table*}[t]
\centering
\scriptsize
\label{tab:training_prompts}
\renewcommand{\arraystretch}{1.15}
\begin{tabular}{p{1cm}|p{2cm}|p{10.8cm}}
\hline
\textbf{Setting} & \textbf{Model} & \textbf{Prompt Format} \\
\hline

Zero-shot & LLaMA-3.2-3B-Instruct & 
\texttt{<|begin\_of\_text|><|start\_header\_id|>system<|end\_header\_id|>} \newline
\texttt{You are a helpful, respectful, and honest assistant. Always answer as helpfully as possible.} \newline
\texttt{<|eot\_id|><|start\_header\_id|>user<|end\_header\_id|> \{sample['seekerpost']\} <|eot\_id|>} \newline
\texttt{<|start\_header\_id|>assistant<|end\_header\_id|>} \\

\hline

Instruction-tuned & Mistral-7B-v0.1 &
\texttt{Instruction: You are a helpful, respectful, and honest assistant.} \newline
\texttt{Input text: \{sample['seeker\_post']\}} \newline
\texttt{Response: \{sample['response\_post']\}} \newline
\texttt{End} \\

\hline

Emotion fine-tuning & LLaMA-3.2-3B-Instruct &
\texttt{<|begin\_of\_text|><|start\_header\_id|>system<|end\_header\_id|>} \newline
\texttt{You are a helpful, respectful, and honest assistant. Always answer as helpfully.} \newline
\texttt{<|eot\_id|><|start\_header\_id|>user<|end\_header\_id|> \{sample['seeker\_post']\} <|eot\_id|>} \newline
\texttt{<|start\_header\_id|>assistant<|end\_header\_id|> \{sample['response\_post']\} <|eot\_id|>} \\

\hline
\end{tabular}
\caption{Prompt templates used for training on the hate/neutral task. Prompts are adapted based on the model setting.}
\label{tab:prompts}
\end{table*}

\section{Weak-supervision training}
\label{app:weaktrainingMacnamer}
As a baseline, we implement a multi-label SVM and LR classifier using a One-vs-Rest LinearSVC with TF-IDF features, where key hyperparameters (e.g., \texttt{C}, \texttt{max\_features}, and \texttt{ngram\_range}) are optimized via grid search with 3-fold cross-validation. 
Also, we fine-tune \texttt{Mistral-7B-Instruct-v0.1} for multi-label classification using QLoRA with a custom classification head, applying LoRA adapters (\texttt{r=16}, \texttt{alpha=32}, dropout=0.1) and optimizing with \texttt{AdamW} and cosine learning rate schedule over three epochs.

For BiLSTM, the model uses a token-level embedding layer followed by a bidirectional LSTM (BiLSTM) for the contextual dependencies in both directions. The final hidden states from both directions are concatenated and passed through a fully connected layer to produce multi-label logits for the four categories. As for the ensemble, the model architecture consists of a pretrained Transformer encoder (DeBERTa or MentalBERT) followed by layer normalization, dropout, and a two-layer feedforward classifier with ReLU activation for multi-label prediction, Table~\ref{tab:ensemble_hyperparams}.

\begin{table}[ht]
\centering
\small
\begin{tabular}{|l|l|}
\hline
\textbf{Hyperparameter} & \textbf{Value} \\
\hline
\texttt{max\_features} & 5000 \\
\texttt{max\_len} (sequence length) & 50 \\
\texttt{batch\_size} & 16 \\
\texttt{optimizer} & Adam \\
\texttt{learning\_rate} & 1e-3 \\
\texttt{loss\_function} & BCELogitsLoss \\
\texttt{epochs} & 5 \\
\hline
\end{tabular}
\label{tab:bilstm_hyperparams}
\caption{BiLSTM model and training hyperparameters.}
\end{table}

\section{Validation of the analysis}
\label{app:validitionAnalysis}
 We conducted Chi-squared tests (and Fisher's exact test when expected counts were low) to assess whether the use of incongruant/misguided positivity labels differed significantly between Severe and Mild concern contexts. As shown in Table~\ref {tab:misguided-chi}, there is a substantial variation by both label and source. For example, humans used the \textit{Minimizing} label significantly more in mild contexts (56.8\%) compared to severe ones (26.75\%), whereas LLMs showed the opposite trend. Similarly, the \textit{None} category—indicating no misguided positivity—was significantly more common in severe human responses (64.16\%), suggesting greater emotional sensitivity. Overall, 7 out of 8 comparisons showed statistically significant differences ($p < .05$), highlighting distinct behavioral patterns in how humans and LLMs respond to varying emotional intensities.

To further validate the significance of the reframing strategies of the comparison, we provide a detailed chi-squared significance test in Table~\ref{tab:reframing-significance-effectsize}. 
The chi-squared significance tests revealed notable differences in reframing strategies across concern types. Specifically, LLM responses showed a strong shift from \textit{None} to more supportive strategies like \textit{Impermanence} and \textit{Growth} when addressing mild concerns, suggesting a potential overcompensation or context miscalibration. In contrast, human responses significantly favored \textit{Relatability}, \textit{Optimism}, and \textit{Impermanence} in mild contexts, while strategies like \textit{None} were more prevalent in severe concerns. These disparities highlight how both source and concern severity shape the affective framing of support, with LLMs and humans exhibiting distinct patterns of alignment with emotional intensity. Moreover, the table shows Cramér’s V test to guage how much the use of each strategy differs based on the severity of user concerns. Large values suggest context-sensitive behavior (or overcompensation), while small or zero values suggest context-independent usage or stable framing. \noindent
Cramér’s V values revealed large effect sizes ($V > 0.5$) for LLM strategies such as \textit{None} and \textit{Impermanence}, indicating strong shifts in reframing behavior between severe and mild concerns. In contrast, human responses showed small to moderate effect sizes ($V \approx 0.14$–$0.26$) for strategies like \textit{Relatability}, \textit{Optimism}, and \textit{Impermanence}, suggesting more nuanced adjustments. These findings highlight that while both LLMs and humans adapt to concern severity, LLMs exhibit sharper, more polarized changes in strategy use.

\begin{table}[ht]
\centering
\small
\begin{tabular}{l l l l}
\toprule
\textbf{Label} & \textbf{Source} & \textbf{Test} & \textbf{p-value} \\
\midrule
Dismissive     & Human  & Chi2   & 0.025$^*$ \\
Dismissive     & LLM    & Chi2   & $< 10^{-7}$$^*$ \\
Minimizing     & Human  & Chi2   & $< 10^{-16}$$^*$ \\
Minimizing     & LLM    & Chi2   & $< 10^{-4}$$^*$ \\
None           & Human  & Chi2   & $< 10^{-50}$$^*$ \\
None           & LLM    & Chi2   & $< 10^{-11}$$^*$ \\
Unrealistic    & Human  & Chi2   & $< 10^{-14}$$^*$ \\
Unrealistic    & LLM    & Fisher & 0.446 \\
\bottomrule
\end{tabular}
\caption{Chi-squared (and Fisher) test results comparing the distribution of misguided positivity labels between Severe and Mild concerns, across human and LLM sources. Asterisks ($^*$) indicate significance at $p < .05$.}
\label{tab:misguided-chi}
\end{table}

\begin{table}[ht]
\centering
\small
\begin{tabular}{lccc}
\toprule
\textbf{Reframing} & \textbf{Source} & \textbf{p} & \textbf{Cramér’s} \\
\midrule
None               & LLM    & $< 10^{-58}$\textsuperscript{*} & 0.578 \\
Impermanence       & LLM    & $< 10^{-58}$\textsuperscript{*} & 0.577 \\
Relatability       & Human  & $3.1 \times 10^{-13}$\textsuperscript{*} & 0.263 \\
None               & Human  & $8.9 \times 10^{-10}$\textsuperscript{*} & 0.221 \\
Optimism           & Human  & $4.5 \times 10^{-5}$\textsuperscript{*} & 0.147 \\
Impermanence       & Human  & $6.3 \times 10^{-5}$\textsuperscript{*} & 0.144 \\
Growth             & LLM    & $7.5 \times 10^{-4}$\textsuperscript{*} & 0.120 \\
Thankfulness       & LLM    & $0.033$\textsuperscript{*} & 0.076 \\
Thankfulness       & Human  & $0.145$ & 0.053 \\
Relatability       & LLM    & $0.951$ & 0.002 \\
Optimism           & LLM    & $1.000$ & 0.000 \\
Growth             & Human  & $1.000$ & 0.000 \\
\bottomrule
\end{tabular}
\caption{Chi-squared test results and effect sizes (Cramér’s V) for reframing strategies by source. Asterisks (\textsuperscript{*}) indicate significance at $p < .05$. Effect sizes: small ($\sim$0.1), medium ($\sim$0.3), large ($\ge$0.5).}
\label{tab:reframing-significance-effectsize}
\end{table}
\begin{table}[htbp]
\centering

\begin{tabular}{lc}
\toprule
\textbf{Relief Category} & \textbf{p-value} \\
\midrule
LLM Better (A > B)       & $6.44 \times 10^{-35}$\textsuperscript{***} \\
Human Better (B > A)     & $1.09 \times 10^{-99}$\textsuperscript{***} \\
Equal Relief             & $1.20 \times 10^{-3}$\textsuperscript{**} \\
No Relief                & $1.08 \times 10^{-9}$\textsuperscript{***} \\
\bottomrule
\end{tabular}

\caption{
Statistical comparison of emotional relief preferences between mild (Advice) and severe (Grief) concern contexts using Chi-squared tests. The p-values indicate whether the frequency of annotator judgments for each relief category differs significantly across contexts. Stars denote significance levels.
 The null hypothesis assumes no difference in relief preference distribution between concern severities. \textsuperscript{**} $p < .05$, \textsuperscript{***} $p < .005$}
\label{tab:chi_significance}

\end{table}

\section{\added{Detailed per-model error analysis}}
\label{app:detail_error}
\added{As shown in Table~\ref{tab:fpr_fnr_misguided}, we carried out a detailed analysis of how the top three models behave. Of all the models tested, the ensemble model performs best. It avoids the over-prediction problems seen in DeBERTa and BERT and achieves a better balance between false positives and false negatives across different labels. For the Minimizing and None labels, all models have very low false negative rates, showing almost perfect recall. However, they also have very high false positive rates, up to 1.000 for DeBERTa and BERT. This means these models often over-predict positivity-related labels and cannot tell the difference between misguided positivity and truly supportive responses.
The Dismissive label shows the opposite trend, with low false positive rates but high false negative rates for all models. This suggests the models often miss dismissive responses, likely because dismissiveness is subtle and depends on context, making it hard to detect from surface-level language. For the Unrealistic label, results are more balanced, especially for the ensemble model, which has a low false positive rate and a moderate false negative rate. This suggests that clear signs of over-optimism or implausibility are easier for models to spot, though they still miss some cases. }

\begin{table}[t]
\centering

\setlength{\tabcolsep}{9pt}
\renewcommand{\arraystretch}{1.15}
\begin{adjustbox}{max width=\linewidth}
\small
\begin{tabular}{llcc}
\toprule
\textbf{Model} & \textbf{Label} & \textbf{FPR} & \textbf{FNR} \\
\midrule
\multirow{4}{*}{\textbf{Ensemble}}
& \Min   & 0.889 & 0.019 \\
& \NoneL & 0.917 & 0.019 \\
& \Dis   & 0.164 & 0.600 \\
& \Unr   & 0.018 & 0.500 \\
\midrule
\multirow{4}{*}{\textbf{DeBERTa}}
& \Min   & \textbf{1.000} & 0.000 \\
& \NoneL & \textbf{1.000} & 0.000 \\
& \Dis   & 0.336 & 0.467 \\
& \Unr   & 0.065 & 0.350 \\
\midrule
\multirow{4}{*}{\textbf{BERT}}
& \Min   & \textbf{1.000} & 0.000 \\
& \NoneL & \textbf{1.000} & 0.000 \\
& \Dis   & 0.261 & 0.633 \\
& \Unr   & 0.061 & 0.500 \\
\bottomrule
\end{tabular}
\end{adjustbox}

\vspace{2pt}
\footnotesize
\textbf{Labels:} \Min{} = Minimizing, \NoneL{} = No Misguided Positivity,
\Dis{} = Dismissive, \Unr{} = Unrealistic.
\caption{\added{False Positive Rate (FPR) and False Negative Rate (FNR) for misguided positivity detection.
Worst values (degenerate behavior) are shown in \textbf{bold}.}}
\label{tab:fpr_fnr_misguided}
\end{table}

\end{document}